\documentclass[letterpaper, 10 pt, conference]{ieeeconf}

\IEEEoverridecommandlockouts                              

\usepackage{cite}
\usepackage{amsmath,amssymb,amsfonts}
\usepackage{algorithmic}
\usepackage{graphicx}
\usepackage{float}
\usepackage{textcomp}
\usepackage{xcolor}
\usepackage[hidelinks]{hyperref}
\usepackage{amsmath}
\usepackage{subcaption}
\usepackage{booktabs}

\newcommand{\meanstd}[2]{%
    \ensuremath{#1\,\mbox{\footnotesize$\pm\,#2$}}%
}

\newcommand{\bestmeanstd}[2]{%
    \ensuremath{\mathbf{#1}\,%
    \mbox{\footnotesize$\boldsymbol{\pm}\,\mathbf{#2}$}}%
}

\title{\LARGE \bf
Energy-Efficient Gait Adaptation via Hierarchical Reinforcement Learning for Quadrupedal Locomotion Across Diverse Terrains
}

\author{Ammar Issa$^{*}{†}$ Anubhav Singh$^{*}$ Anton Tsaritsin$^{*}$ Sergey Kolyubin\thanks{*These authors contributed equally to this work.}\thanks{All authors are affiliated with the Faculty of Control Systems and Robotics, ITMO University, and the Biomechatronics and Energy-Efficient Robotics Lab (BE2R), St.~Petersburg, Russia (e-mail: \{issa.ammar, anubhav.s, avtsaritsin, s.kolyubin\}@itmo.ru).}}

\begin{document}

\maketitle
\thispagestyle{empty}
\pagestyle{empty}


\begin{abstract}

While energy efficiency is a critical objective for legged-robot locomotion control, achieving low energy consumption while maintaining robust performance across different velocity ranges and terrain conditions remains a key challenge. This is particularly true for end-to-end RL policies, where gait generation, motion execution, and energy optimization are tightly coupled, leading to high sensitivity to reward design. In this work, we propose a hierarchical reinforcement learning (HRL) framework that separates a high-frequency policy for stable and robust joint-level motion execution from low-frequency gait adaptation that explicitly minimizes the cost of transport (CoT). The three-stage Isaac-based training procedure enables zero-shot sim-to-real transfer with improved tracking accuracy, robustness, and energy efficiency. The learned hierarchy exhibits automatic speed-dependent gait adaptation, transitioning from pacing at low speeds to trotting at higher speeds. We validate the proposed approach in simulation against representative single-policy and hierarchical locomotion baselines, demonstrating reduced CoT over a broad range of commanded velocities, while maintaining robust locomotion across flat, uneven rough, and inclined terrains. We further demonstrate its practical feasibility through zero-shot deployment on a physical Unitree AlienGo quadruped.


\end{abstract}

\section{Introduction}

Energy efficiency is essential for quadruped robots operating in
long-duration and energy-constrained tasks. Their nonlinear
dynamics, intermittent contacts, and speed-dependent gait
energetics make it difficult to achieve both stable locomotion and
low energy consumption. Biological studies show that different
gaits become energetically favorable at different speeds
\cite{hoyt1981gait}, motivating controllers that adapt gait rather
than rely on a single locomotion pattern.

Model-based methods use model predictive control (MPC), trajectory optimization, and quadratic programming (QP)-based control to generate dynamically feasible and energy-conscious
motions \cite{di2018dynamic,shamraev2022bioinspired,
alqaham2024energy,alqaham202516}. Bio-inspired approaches exploit
passive dynamics or central pattern generators (CPGs) to obtain interpretable gait structures
\cite{masuda2021brainless,arena2021energy}, with recent methods
using reinforcement learning (RL) to adapt oscillator parameters and
generate multiple gaits and transitions
\cite{bellegarda2022cpg,humphreys2024learning,
bellegarda2024allgaitslearningquadrupedgaits}. However, these
approaches often depend on accurate models, predefined motion
structures, or hand-designed oscillator representations.

\begin{figure}[t]
    \centering
    \includegraphics[width=0.9\columnwidth]{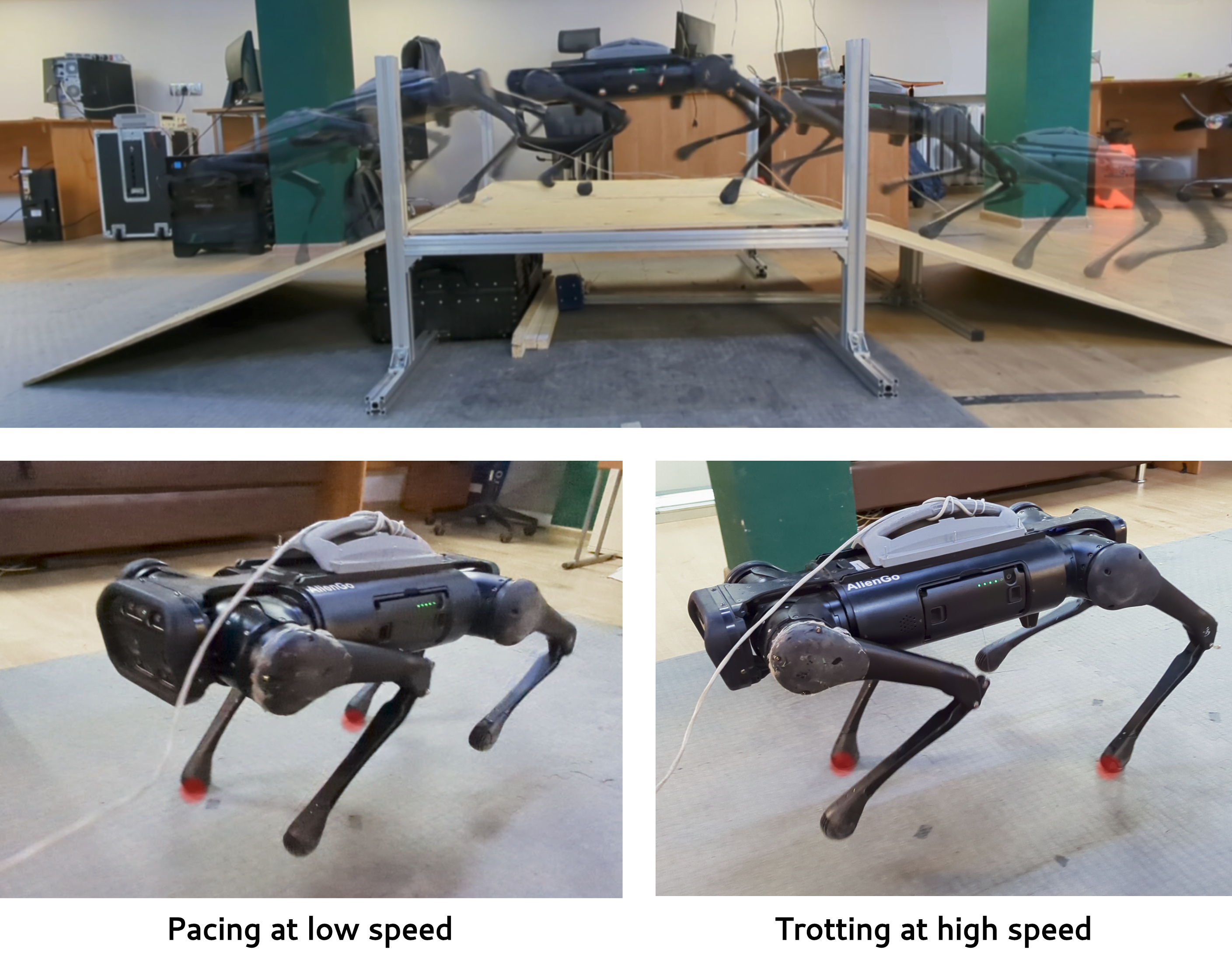}
    \caption{Stable locomotion and slope traversal, with
    speed-dependent gait modulation from pacing at lower speeds to
    trotting at higher speeds.}
    \label{fig:real_teas}
\end{figure}

RL has enabled robust quadrupedal locomotion
without explicit dynamics models
\cite{lee2020learning,rudin2022learning}. Such policies are
typically trained using algorithms such as proximal policy optimization (PPO) \cite{schulman2017proximal}, with tracking, stability, imitation,
and energy objectives combined through reward design
\cite{peng2018deepmimic,margolis2023walk}. Energy-aware methods
penalize power or CoT and can produce efficient, speed-dependent
gait transitions
\cite{hao2024energyefficient,santos2025energy,
liang2025adaptive,wang2025multigait}. Nevertheless, end-to-end
policies jointly optimize gait generation, stabilization, tracking,
and energy efficiency, making them sensitive to reward weighting
and reducing interpretability.

Hierarchical methods address this coupling by separating
gait-level decisions from low-level motor control
\cite{jain2019hierarchical}. Existing approaches generate
CPG-based gait commands \cite{learnmultihrl211204741}, combine
learned gait parameters with MPC \cite{yang2022fast}, or switch
between pretrained energy-efficient gait policies
\cite{stamatopoulou2025egait}. While effective, many still rely on
discrete switching, predefined gait libraries, or model-based
low-level controllers.

In this work, we propose a fully learning-based hierarchical
reinforcement learning framework for energy-efficient quadruped
locomotion across different speeds and terrains. A high-frequency
low-level policy performs velocity tracking and stabilization,
while a slower high-level policy continuously modulates
interpretable gait parameters. Energy efficiency is incorporated
directly into the high-level objective through CoT-related
optimization, enabling adaptive gait modulation without predefined
gait libraries or manually designed transition rules.
Representative locomotion behaviors, including flat-ground
walking, slope traversal, and speed-dependent gait modulation,
are shown in Fig.~\ref{fig:real_teas}.


The main contributions of this work are:
\begin{itemize}
    \item A hierarchical RL framework that separates low-frequency gait adaptation
    from high-frequency velocity tracking and stabilization, followed
    by post-training finetuning.

    \item A high-level objective that explicitly incorporates
    CoT-related energy optimization for continuous gait-parameter
    modulation.

    \item Evaluation of tracking and CoT across different speeds and
    terrains, together with zero-shot deployment on a physical Unitree AlienGo.
\end{itemize}

\section{Hierarchical Reinforcement Learning Framework}

\subsection{Problem Formulation}

Quadruped locomotion is formulated as a Markov decision process
(MDP) defined by the tuple
$(\mathcal{S},\mathcal{A},\mathcal{P},r)$, representing the
state space, action space, transition dynamics, and reward
function, respectively. To account for different temporal
scales of gait adaptation and joint-level control, the policy is
decomposed into a low-frequency high-level (HL) policy
$\pi_{HL}$ and a high-frequency low-level (LL) policy
$\pi_{LL}$.

Let $H$ denote the decimation factor between the HL and LL policies, i.e. for each HL command $g_k$, the LL controller executes $H$ consecutive control steps:
\begin{equation}
\mathbf{a}_t = \pi_{LL}(o_t^{LL}, g_k; \phi_{ll}),
\quad t \in \{kH, \ldots, (k+1)H - 1\},
\label{eq:ll_action}
\end{equation}
where $o_t^{LL}$ denotes the LL observation at time step $t$, which includes joint states, base orientation information, and phase-based timing signals, and $\phi_{ll}$ represents the parameters of the LL policy network.

\subsection{Hierarchical Control Architecture}

\begin{figure}[t]
\centering
\includegraphics[width=0.98\columnwidth]{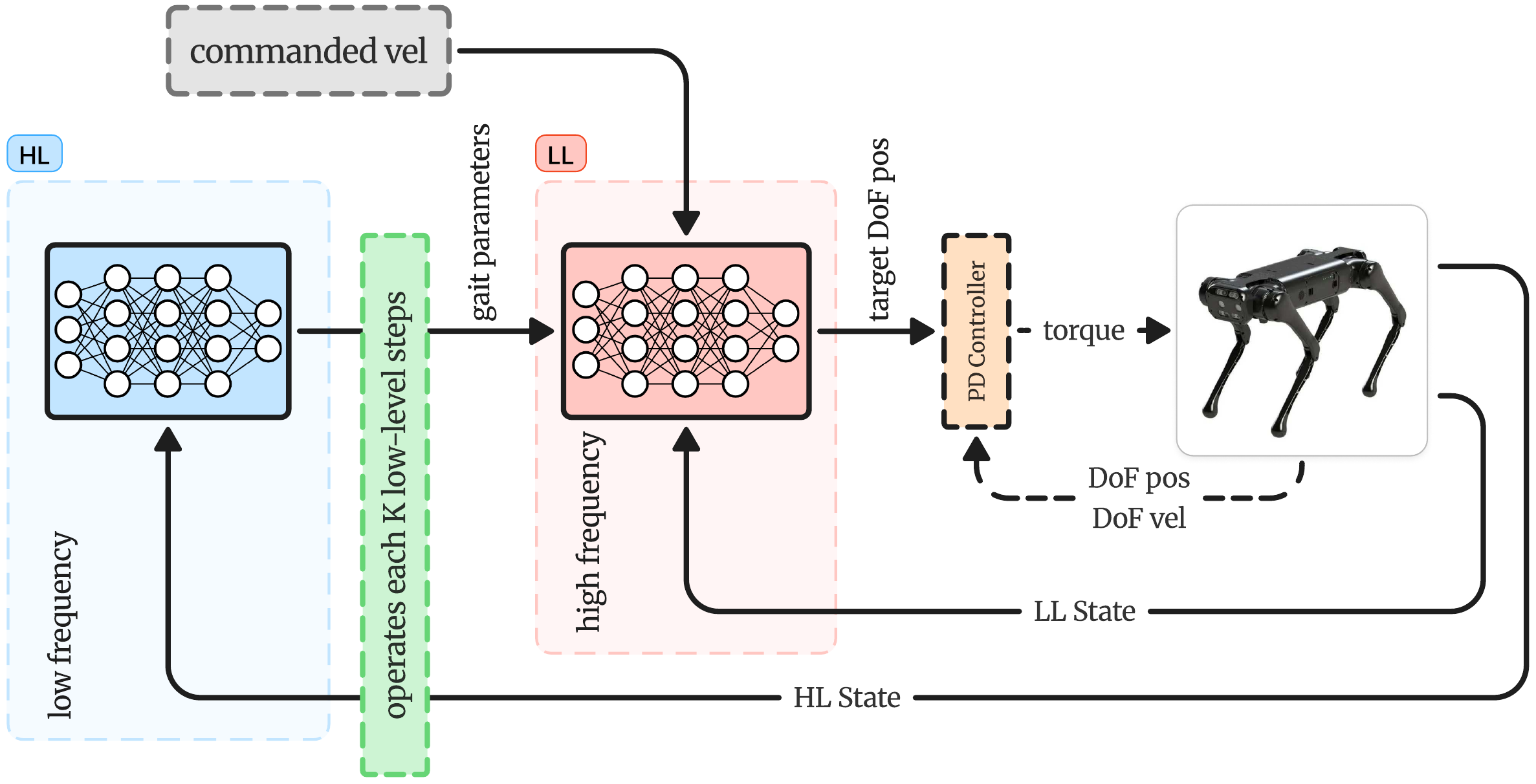}
\caption{Overview of the hierarchical locomotion control framework. The high-level policy operates at a lower frequency to modulate gait parameters, while the low-level controller performs high-frequency motion execution.}
\label{fig:hrl}
\end{figure}

The proposed architecture consists of a HL gait generation policy and a LL locomotion controller operating at different temporal scales, as illustrated in Fig.~\ref{fig:hrl}. The HL policy is evaluated at a lower frequency and outputs a structured gait parameter vector $g_k$. The LL controller operates at a higher frequency and computes joint-level commands according to Eq.~(\ref{eq:ll_action}).

At each HL decision step, the gait parameters remain fixed for $H$ low-level control steps, defining a control horizon over which locomotion behavior unfolds. After this execution window, aggregated locomotion statistics are computed and used to update the HL policy.

The LL policy is trained independently for robust velocity tracking and dynamic stabilization and is kept frozen during HL training. This design isolates contact-rich dynamics within the LL controller and allows the HL policy to focus on long-horizon gait modulation and energy optimization.

\subsection{High-Level Gait Parameterization}


Formally, the HL policy is defined as a stochastic policy
\begin{equation}
g_k \sim \pi_{HL}( \cdot \mid s_k^{HL}; \phi_{hl}),
\label{eq:hl_action}
\end{equation}
where $g_k \in \mathbb{R}^d$ denotes the gait command at HL timestep $k$, $s_k^{HL}$ is the HL state, and $\phi_{hl}$ represents the parameters of the HL policy network. Each HL action remains fixed for $H$ consecutive LL control steps and is provided as a conditioning input to the LL controller.

\paragraph{High-Level State Space}

The high-level state at decision step $k$ is defined as

\begin{equation}
\mathbf{s}_k^{HL}
=
\left[
\bar{v}_{x,k},\;
\bar{v}_{y,k},\;
\bar{\omega}_{z,k},\;
\mathbf{g}_{\mathrm{proj},k}
\right]
\in \mathbb{R}^{6}.
\end{equation}

where $\bar{v}_{x,k}$ and $\bar{v}_{y,k}$ are the body-frame
linear-velocity components averaged over the LL execution
window associated with HL step $k$, $\bar{\omega}_{z,k}$
is the yaw angular velocity averaged over the same window, and
$\mathbf{g}_{\mathrm{proj},k} \in \mathbb{R}^{3}$ is the
instantaneous gravity vector expressed in the body frame at the
end of that window. By summarizing motion over the LL control horizon associated with each HL action, the HL policy reasons over aggregated locomotion behavior rather than instantaneous joint-level states. This temporal abstraction enables long-horizon gait adaptation, while short-term stabilization is handled by the LL controller.


Commanded velocities are not explicitly included in $s_k^{HL}$. Instead, the HL policy adapts gait parameters based on realized motion feedback, encouraging modulation according to actual locomotion performance rather than desired commands. Meanwhile, an asymmetric actor-critic architecture was employed \cite{pinto2017asymmetric}, where the commanded gait parameters are provided only to the critic network to enable better value estimation, since the HL state alone is insufficient for accurately estimating CoT-related signals. By operating on aggregated global motion and orientation signals, the HL policy focuses on long-horizon gait adaptation and energy regulation, while short-term stabilization and joint-level tracking are handled by the LL control policy.

\paragraph{High-Level Action Space}


\begin{figure*}[t]
    \centering
    \includegraphics[width=\textwidth]{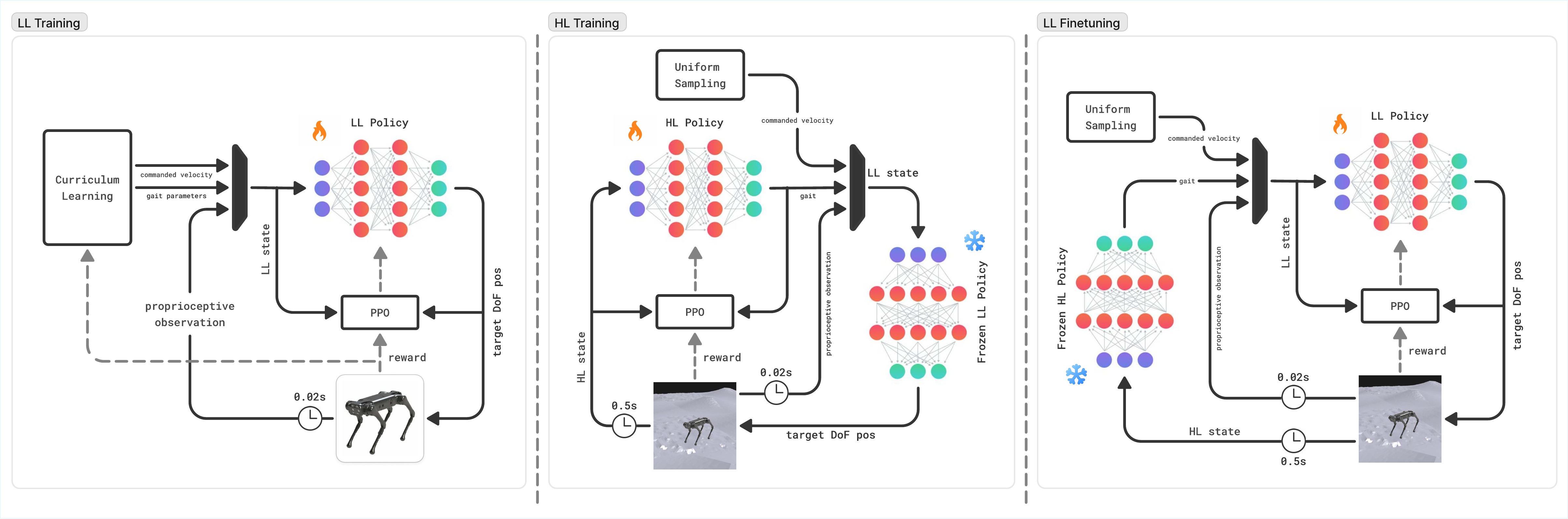}
    \caption{Three-stage training pipeline: LL policy
    pretraining, energy-aware HL policy training with the LL
    policy frozen, and LL policy finetuning with the HL policy
    frozen.}
    \label{fig:train_pipeline}
\end{figure*}

The HL action $g_k \in \mathbb{R}^{11}$ parametrizes the gait
executed by the LL controller and is defined as
\begin{equation}
g_k = \left[ \theta_1, \theta_2, \theta_3, f_{\mathrm{gait}}, h_z, \theta_x, \theta_y, s_x, s_y, h_{f,z}, d \right],
\end{equation}
where $(\theta_1,\theta_2,\theta_3)$ are inter-leg timing offsets
that determine the phase relationships among the four legs,
$f_{\mathrm{gait}}$ is the stepping frequency, $h_z$ is the body
height, $\theta_x$ and $\theta_y$ are the body roll and pitch,
respectively, $s_x$ and $s_y$ are the stance length and width,
$h_{f,z}$ is the foot-swing height, and $d$ is the stance duration.

These parameters define the behavioral reference used by the LL policy, modulating gait structure without directly altering joint-level motion execution. In particular, the timing offsets implicitly encode canonical quadruped gaits such as pronking, trotting, pacing, or bounding, while allowing smooth interpolation through continuous modulation.

By restricting the HL policy to operate on structured gait parameters and reusing a pretrained LL control policy for joint-based execution, the hierarchical formulation decouples long-horizon energy optimization from short-horizon dynamic stabilization. This design enables adaptive and energy-aware gait modulation while preserving robust locomotion performance across varying operating conditions.

\paragraph{High-Level Reward}

The HL reward combines energy efficiency, gait smoothness,
and accumulated LL performance:
\begin{equation}
r_k^{HL}
=
r_{\mathrm{energy}}
-0.1\|g_k-g_{k-1}\|^2
+0.002\sum_{t=kH}^{(k+1)H-1}r_t^{LL}.
\label{eq:hl_reward}
\end{equation}
The energy term is inspired by \cite{liang2025adaptive}, and defined as:
\begin{equation}
r_{\mathrm{energy}}=
\exp\!\left(
-\frac{
\sum_{t=kH}^{(k+1)H-1}
\sum_{i=1}^{12}|P_t^i|
}{
\sigma_{\mathrm{en},x}|\bar v_x|
+\sigma_{\mathrm{en},z}|\bar\omega_z|
}
\right),
\label{eq:energy_reward}
\end{equation}
where $P_t^i=\tau_t^i\dot q_t^i$ is the mechanical power of
joint $i$. The LL reward prevents energy minimization through
unstable or poorly tracked motion, such as crouching behavior, while the smoothness term penalizes abrupt changes in gait commands. With
$\Delta t_{LL}=0.02$\,s and $\Delta t_{HL}=0.5$\,s, each HL
reward is accumulated over $H=25$ LL steps.

\subsection{Low-Level Locomotion Policy}




\paragraph{Low-Level State Space}

The LL observation is
\begin{equation}
s_t^{LL} =
[
\mathbf{g}_{\mathrm{proj},t},
\mathbf{c}_t,
\mathbf{q}_t-\mathbf{q}^{0},
\dot{\mathbf{q}}_t,
\mathbf{a}_{t-1},
\mathbf{clk}_t
]\in \mathbb{R}^{58},
\end{equation}
where $\mathbf{c}_t$ contains the commanded base velocities
and gait parameters, $\mathbf{q}_t-\mathbf{q}^{0}$ and
$\dot{\mathbf{q}}_t$ are the joint position offsets and
velocities, and $\mathbf{clk}_t$ contains four phase-clock
signals. The policy receives a history of $H_h=5$ observations,
$h_t^{LL}=[s_{t-H_h+1}^{LL},\ldots,s_t^{LL}]$.

\begin{figure*}[t]
    \centering
    \includegraphics[width=\textwidth]{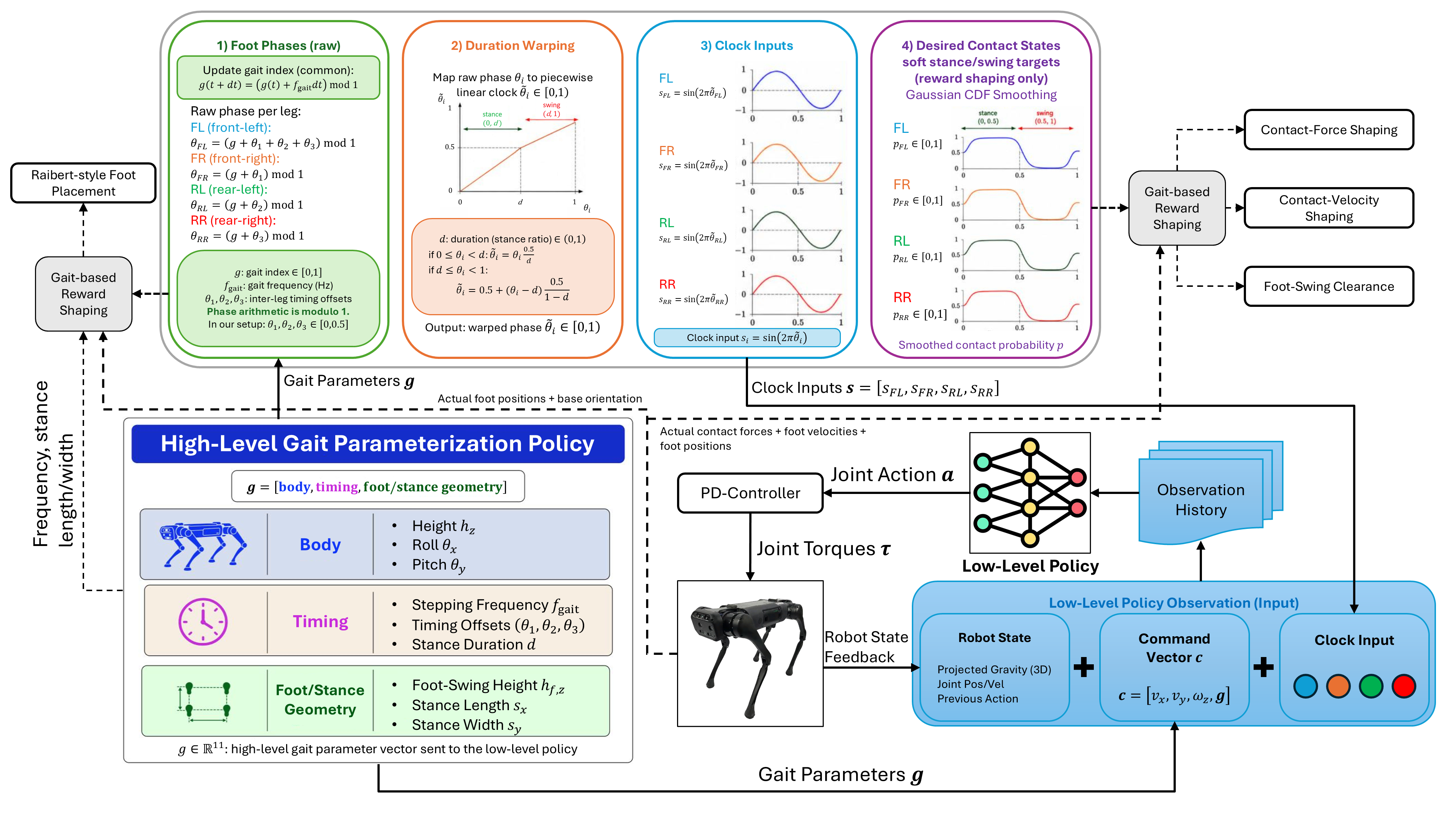}
    \caption{Processing of commanded gait parameters into leg-wise phases, clock inputs, and training-time contact targets.}
    \label{fig:gait_parameterization}
\end{figure*}

Figure~\ref{fig:gait_parameterization} illustrates how the commanded gait parameters are incorporated into the LL policy. The timing parameters define leg-wise phase indices, which are warped using the stance duration parameter and transformed into sinusoidal clock inputs.
These clock inputs, together with the gait command vector and robot-state observations, condition the LL policy. The desired contact states are phase-derived soft stance/swing targets used only for reward shaping during
training; they are not policy inputs and do not directly prescribe Cartesian foot trajectories. During LL pretraining, the gait-command vector $g$ is sampled by the curriculum, whereas after introduction of the hierarchical controller it is generated by the HL policy.

\paragraph{Low-Level Action Space}

The LL action $\mathbf{a}_t\in\mathbb{R}^{12}$ contains desired
joint positions tracked by a joint-level PD controller.

\paragraph{Low-Level Reward}

The LL reward follows the formulation used in \cite{margolis2023walk}. It is divided into three groups: (i) task reward for velocity tracking, (ii) augmented auxiliary rewards that enforce the commanded gait behavior, and (iii) fixed auxiliary rewards that encourage stable and smooth locomotion. 
The augmented auxiliary reward explicitly associate the gait parameters with the desired locomotion behavior. Body roll $\theta_x$ and pitch $\theta_y$ are enforced through an orientation-tracking term $r_{\mathrm{ori}}$, foot-swing height $h_{f,z}$ through $r_{\mathrm{clearance}}$, and body height $h_z$ through $r_{\mathrm{jump}}$. The timing offsets $(\theta_1,\theta_2,\theta_3)$, stepping frequency $f_{\mathrm{gait}}$, and stance duration $d$ define the desired per-foot contact schedule, which is enforced through the
contact-force and contact-velocity terms $r_{\mathrm{contact\_force}}$ and
$r_{\mathrm{contact\_velocity}}$. Finally, a Raibert-style foot-placement term $r_{\mathrm{raibert}}$ encourages foothold placement consistent with the commanded gait and stance geometry~\cite{raibert1986legged}. The augmented auxiliary rewards are summarized in Table~\ref{tab:ll_aug_rewards}, while the remaining reward terms follow~\cite{margolis2023walk}. No additional energy-specific shaping is introduced at the LL stage.

\begin{table}[t]
    \centering
    \caption{Augmented auxiliary rewards used to enforce the commanded gait parameters in the LL policy.}
    \label{tab:ll_aug_rewards}
    \renewcommand{\arraystretch}{1.25}
    \setlength{\tabcolsep}{3pt}
    \resizebox{\columnwidth}{!}{%
    \begin{tabular}{lll}
        \toprule
        \textbf{Reward} & \textbf{Primary gait parameter} & \textbf{Formulation} \\
        \midrule
        
        $r_{\mathrm{ori}}$
        &
        $\theta_x,\theta_y$ 
        &
        $\left\|
        \mathbf{g}^{\mathrm{cmd}}_{\text{proj},\{x,y\}}
        -\mathbf{g}_{\text{proj},\{x,y\}}
        \right\|^2$
        \\
        
        $r_{\mathrm{clearance}}$
        &
        $h_{f,z}$ 
        &
        $\displaystyle
        \sum_{i=1}^{4}
        \left(
        h^i_{f,z}
        -h^{\mathrm{target},i}_{f,z}
        \right)^2$
        \\
        
        $r_{\mathrm{jump}}$
        &
        $h_z$ 
        &
        $\left|
        h_z^{\mathrm{cmd}}-h^\text{body}_z+h_z^{\mathrm{ref}}
        \right|$
        \\
        
        $r_{\mathrm{contact\_force}}$
        &
        \begin{tabular}[c]{@{}l@{}}
        $d$ 
        \end{tabular}
        &
        $\displaystyle
        \sum_{i=1}^{4}
        (1-d^i)
        \left[
        1-\exp\left(-\frac{F_i^2}{\sigma_F}\right)
        \right]$
        \\
        
        $r_{\mathrm{contact\_velocity}}$
        &
        \begin{tabular}[c]{@{}l@{}}
        $d$ 
        \end{tabular}
        &
        $\displaystyle
        \sum_{i=1}^{4}
        d^i
        \left[
        1-\exp\left(-\frac{v_i^2}{\sigma_v}\right)
        \right]$
        \\
        
        $r_{\mathrm{raibert}}$
        &
        \begin{tabular}[c]{@{}l@{}}
        $\theta_1,\theta_2,\theta_3,f_\text{gait},s_x,s_y$ 
        \end{tabular}
        &
        $\displaystyle
        \sum_{i=1}^{4}
        \left\|
        \mathbf{p}^{\mathrm{des},i}_{\mathrm{foot}}
        -\mathbf{p}^i_{\mathrm{foot}}
        \right\|^2$
        \\
        
        \bottomrule
    \end{tabular}%
    }
\end{table}

Here, $\mathbf{g}^{\mathrm{cmd}}_{\mathrm{proj},\{x,y\}}$ denotes the desired projected-gravity components obtained from the commanded body roll and pitch, while $\mathbf{g}_{\mathrm{proj},\{x,y\}}$ denotes the corresponding
measured components. The desired foot-swing height $h^{\mathrm{target},i}_{f,z}$ is a phase-dependent trajectory constructed from the commanded foot-swing height and gait cycle, while other timing parameters indirectly affect its temporal evolution through the gait phase. The body height reference
$h_z^{\mathrm{ref}}$ is set to $0.51$~m for AlienGo and $0.30$~m for Go1.

For the contact-related terms, $F_i$ and $v_i$ denote the contact-force magnitude and foot-velocity magnitude of foot $i$, respectively, while $\sigma_F$ and $\sigma_v$ are scaling parameters controlling the corresponding exponential reward shaping. The desired foothold $\mathbf{p}^{\mathrm{des},i}_{\mathrm{foot}}$ is generated from the commanded gait parameters using the Raibert-style heuristic, whereas $\mathbf{p}^{i}_{\mathrm{foot}}$ denotes the realized foot position. The detailed construction of the gait-phase, foot-swing, contact, and foothold references follows \cite{margolis2023walk}.

\subsection{Training Pipeline}

\begin{figure*}[t]
    \centering
    \includegraphics[width=\textwidth]{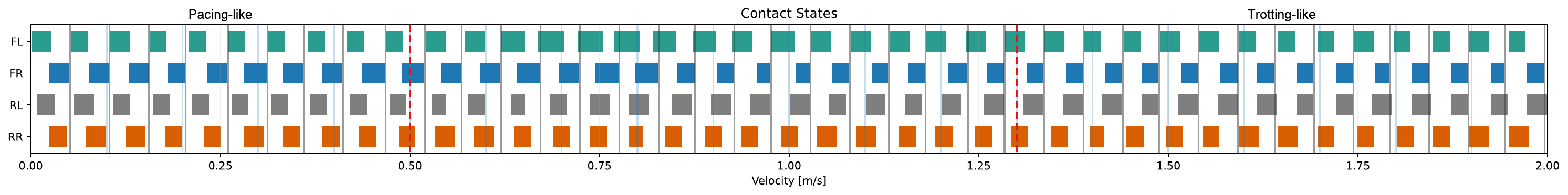}
    \caption{Foot contact state transitions under varying commanded velocities. The high-level policy autonomously and smoothly adapts the gait pattern, exhibiting pacing at low speeds (two-beat walking), a four-beat walking gait at moderate velocities, and trotting at higher speeds.}
    \label{fig:cs}
\end{figure*}

Instead of joint gait adaptation and LL control policies training, we propose a three-stage HRL framework that allows to improve sample efficiency and stabilize training in simulation environment with further zero-shot transfer on a real quadrupedal robot. The full training pipeline is illustrated in Fig.~\ref{fig:train_pipeline}.

\textbf{Stage 1: Low-Level Policy Training.}

The LL policy $\pi_{LL}$ is trained on flat terrain to achieve stable locomotion and accurate tracking of velocity and gait commands. This stage follows the multiplicity of behavior (MoB) framework \cite{margolis2023walk}, where the LL policy is implemented as a conditional policy that receives proprioceptive observations in addition to sampled external velocity and gait commands, and is optimized using RL with curriculum learning to improve robustness and generalization across different locomotion behaviors.

\textbf{Stage 2: High-Level Policy Training.}

After convergence of $\pi_{LL}$, its parameters $\phi_{ll}$ are frozen. The HL policy $\pi_{HL}$ is then introduced and trained on complex terrain, including rough and inclined surfaces, to modulate gait parameters with consideration of energy efficiency. Since $\pi_{LL}$ ensures stable motion execution across different commands, curriculum learning for velocity commands is disabled during this stage, and $\pi_{HL}$ operates at a lower update frequency than the LL policy.

\textbf{Stage 3: Low-Level Policy Finetuning.}

Finally, a short finetuning stage is applied to $\pi_{LL}$ under the complex terrain while keeping the HL policy parameters $\phi_{hl}$ frozen. This stage improves tracking robustness on the new terrain conditions without altering the energy optimization behavior learned at the second stage by the HL policy.

\section{Experimental Results}

\subsection{Experimental Setup}

Experiments were primarily conducted on the real-world Unitree AlienGo quadruped platform to evaluate the proposed HRL framework across velocities and terrain conditions. AlienGo serves as the main experimental platform for both simulation training and real-world validation. To enable direct comparison with prior benchmarks, additional experiments are performed in simulation on the Unitree Go1 platform.


\subsubsection{Simulation Environment and Training Details}

All policies were trained in the NVIDIA Isaac Gym simulation environment \cite{makoviychuk2021isaac} using the PPO algorithm \cite{schulman2017proximal}. 4096 GPU-accelerated parallel environments were employed to enable large-scale data collection and efficient policy optimization \cite{rudin2022learning}. Terrains for the second and third stages of training are generated randomly and include the following types: (i) a flat surface, (ii) uneven rough terrain with heights uniformly sampled between 0 and 10 cm, and (iii) inclined slopes, both upward and downward, with angles randomly selected from 0.2, 0.3, or 0.36 rad. 

The LL locomotion policy was trained following the procedure and hyper-parameters reported in \cite{margolis2023walk}. No modifications were introduced at this stage in order to maintain consistency with prior work. The HL policy was subsequently trained on top of the frozen LL controller to learn energy-aware gait parameter modulation. 

\begin{figure*}[t]
    \centering
    \begin{subfigure}[t]{0.32\textwidth}
    \includegraphics[width=\linewidth]{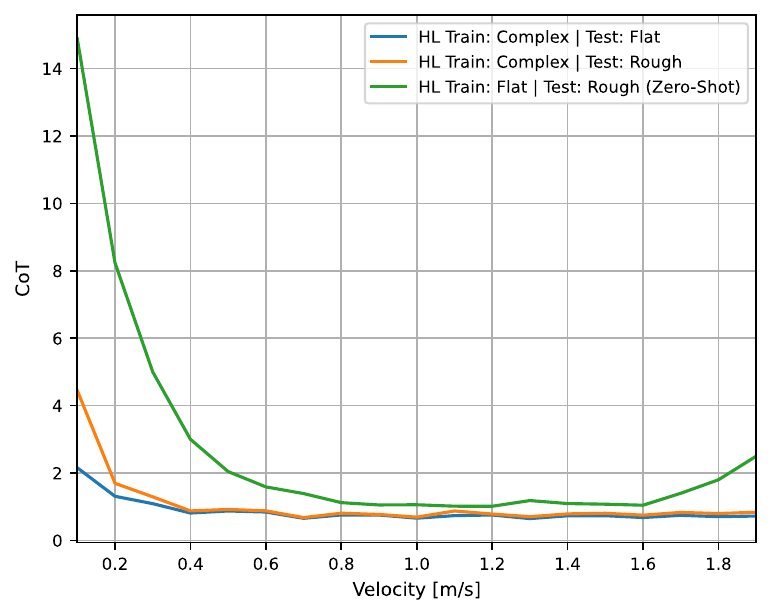}
    \caption{HRL CoT under different terrain-training and deployment conditions.}
    \end{subfigure}
    \begin{subfigure}[t]{0.32\textwidth}
    \includegraphics[width=\linewidth]{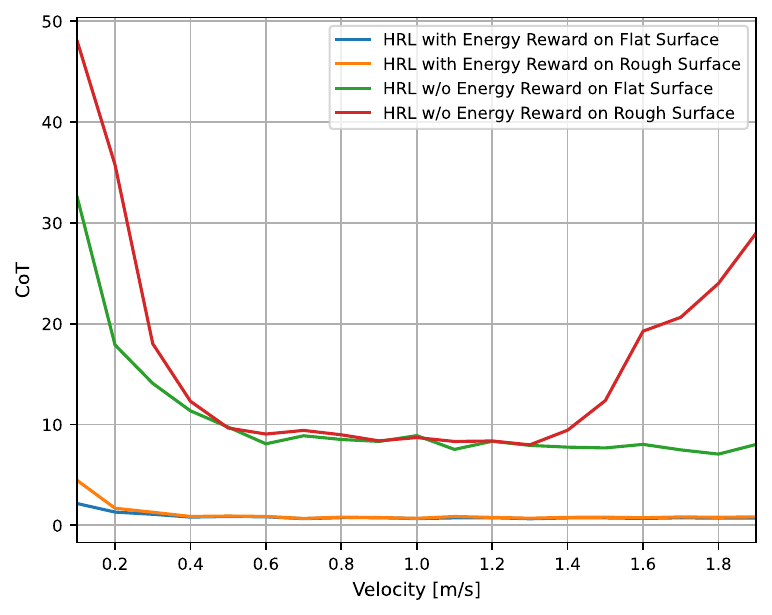}
    \caption{HRL CoT with and without the HL energy reward across different deployment terrains.}
    \end{subfigure}
    \begin{subfigure}[t]{0.32\textwidth}
    \includegraphics[width=\linewidth]{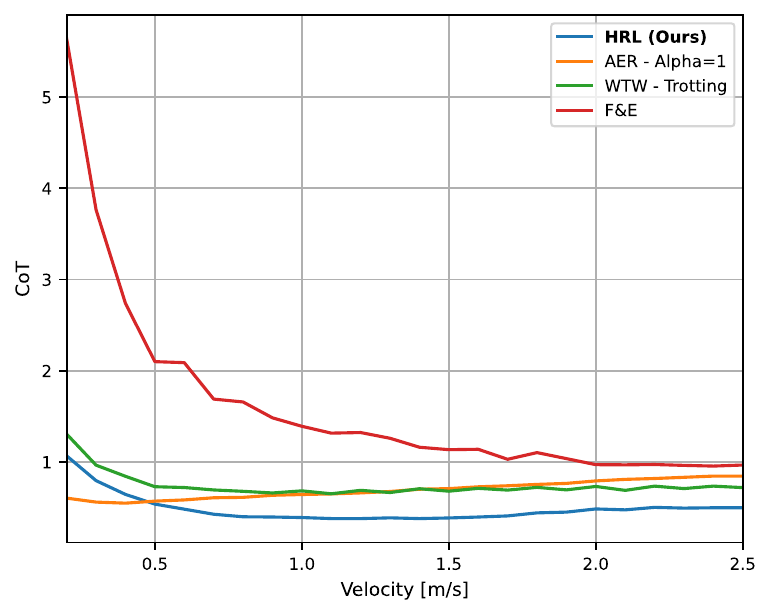}
    \caption{Comparison of HRL CoT with existing locomotion baselines.}
    \end{subfigure}
    \caption{CoT evaluation of the proposed HRL framework. (a) Ablation study on the HL training terrain for the Unitree AlienGo, comparing deployment on flat and rough surfaces, including zero-shot deployment of the flat-trained HL policy on rough terrain. (b) Ablation study on the effect of the HL energy reward for the Unitree AlienGo, comparing CoT with and without the energy reward across different deployment terrains. (c) Comparison of the proposed HRL framework with existing locomotion baselines on the Unitree Go1.}
    \label{fig:cot_all}
\end{figure*}

\subsubsection{Evaluation Protocol}

Policies are evaluated over a range of commanded forward velocities, and performance metrics are computed after an initial transient period to exclude startup effects. Evaluation is conducted on three terrain categories: flat terrain, inclined slopes, and uneven rough terrain. On slopes, both uphill and downhill conditions are considered, and the robot was trained on inclinations up to $20^\circ$. All methods are evaluated under identical terrain configurations and command distributions to ensure a valid comparison. Energy efficiency is quantified using the CoT, defined as:
\begin{equation}
    \text{CoT} = \frac{\sum_{i=1}^{12}{\left|P^i\right|}}{mg\left|v_x\right|}
\end{equation}
where $P^i=\tau_i\dot{q}_i$ denotes the mechanical power at joint $i$, $\tau_i$ and $\dot{q}_i$ denote joint torques and velocities respectively, $m$ is the robot mass, $g$ is gravitational acceleration, and $v_x$ denotes the robot's velocity. In addition to CoT, velocity tracking error is reported to assess locomotion accuracy and robustness.

\begin{table}[t]
\centering
\caption{Robot falling rate in simulation across terrain types and commanded-velocity intervals. Each entry aggregates 500 runs with friction and restitution coefficients uniformly randomized in $[0,1]$.}
\label{tab:simulation_falling_rate}
\resizebox{\columnwidth}{!}{
\begin{tabular}{lcccc}
\toprule
\textbf{Terrain} &
\textbf{$[0.0,0.5]$} &
\textbf{$(0.5,1.0]$} &
\textbf{$(1.0,1.5]$} &
\textbf{$(1.5,2.0]$} \\
\midrule
Flat terrain
& 0.0\% & 0.0\% & 0.0\% & 0.0\% \\

Uneven rough terrain
& 0.0\% & 0.0\% & 0.0\% & 0.6\% \\

Inclined slopes
& 2.4\% & 0.4\% & 0.0\% & 0.2\% \\
\bottomrule
\end{tabular}}
\end{table}

Locomotion robustness was additionally evaluated over 500 simulation runs for each terrain and velocity interval, with friction and restitution coefficients uniformly randomized in $[0,1]$. Aggregated falling rates are reported in Table~\ref{tab:simulation_falling_rate}. The higher low-speed fall rate on slopes likely reflects reduced momentum and longer support phases, limiting recovery from perturbations.





\begin{table*}[t!]
\begingroup

\renewcommand{\meanstd}[2]{%
    \ensuremath{#1\,{\scriptstyle \pm\,#2}}%
}
\renewcommand{\bestmeanstd}[2]{%
    \ensuremath{\mathbf{#1}\,%
    {\scriptstyle\boldsymbol{\pm}\,\mathbf{#2}}}%
}

\centering
\caption{CoT comparison with existing locomotion baselines across
commanded-velocity intervals. Values are reported as mean $\pm$
standard deviation over evaluation runs within each commanded-velocity interval.}
\label{tab:cot_baselines}

\resizebox{\textwidth}{!}{%
\begin{tabular}{lccccc}
    \toprule
    \textbf{Method / $v_x^{\mathrm{cmd}}$ [m/s]}
    & $\mathbf{(0.0,0.5]}$
    & $\mathbf{(0.5,1.0]}$
    & $\mathbf{(1.0,1.5]}$
    & $\mathbf{(1.5,2.0]}$
    & $\mathbf{(2.0,2.5]}$ \\
    \midrule

    \textbf{HRL (Ours)}
    & \meanstd{0.762}{0.198}
    & \bestmeanstd{0.421}{0.034}
    & \bestmeanstd{0.384}{0.004}
    & \bestmeanstd{0.438}{0.031}
    & \bestmeanstd{0.495}{0.010} \\

    AER ($\alpha=1$)
    & \bestmeanstd{0.572}{0.021}
    & \meanstd{0.618}{0.021}
    & \meanstd{0.681}{0.022}
    & \meanstd{0.757}{0.022}
    & \meanstd{0.831}{0.014} \\

    WTW--Trotting
    & \meanstd{0.962}{0.216}
    & \meanstd{0.688}{0.020}
    & \meanstd{0.680}{0.019}
    & \meanstd{0.711}{0.015}
    & \meanstd{0.719}{0.017} \\

    F\&E
    & \meanstd{3.560}{1.338}
    & \meanstd{1.663}{0.240}
    & \meanstd{1.240}{0.078}
    & \meanstd{1.057}{0.059}
    & \meanstd{0.966}{0.006} \\

    \bottomrule
\end{tabular}%
}
\endgroup
\end{table*}




\subsection{Experimental Evaluation}

\subsubsection{Gait Adaptation Behavior}

To analyze the behavior of the HL policy, Fig.~\ref{fig:cs} illustrates the contact states across different commanded velocities. At low speeds, the policy converges to a pacing-like gait with extended stance durations. As velocity increases, the contact pattern gradually shifts toward a trotting gait. This transition emerges without explicit gait switching rules, indicating that the HL policy adapts gait structure as a function of velocity while preserving stable locomotion execution through the LL policy. 

\begin{figure}[H]
    \centering
    \includegraphics[width=\columnwidth]{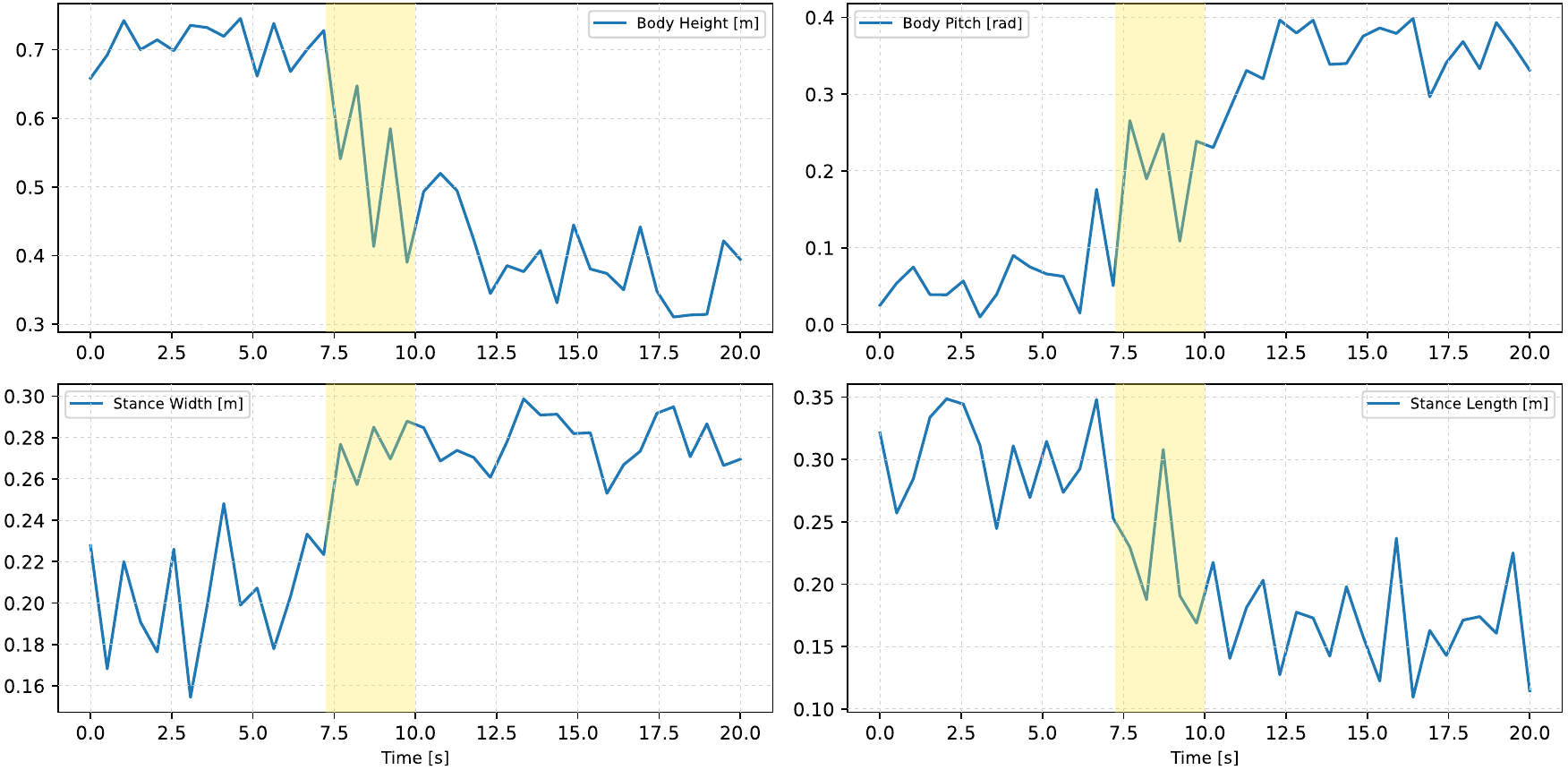}
    \caption{Realized geometrical gait parameters during a representative transition from flat terrain to a $20^\circ$ uphill slope at a fixed commanded forward velocity. The shaded interval marks the terrain-transition phase.}
    \label{fig:geo_gait}
\end{figure}

To complement the speed-dependent contact-timing analysis, Fig. \ref{fig:geo_gait} shows a set of realized geometrical gait parameters during a representative transition from flat terrain to a $20^\circ$ uphill slope at a fixed commanded forward velocity of $1.0$~m/s. The shaded interval denotes the transition phase, during which the robot first encounters the incline and adapts its locomotion. During slope traversal, the realized body height and stance length decrease, while the body pitch and stance width increase, after which the parameters settle around a different operating region.

Together, Figs.~\ref{fig:cs} and \ref{fig:geo_gait} demonstrate that the proposed hierarchy adapts the timing-related gait parameters---the stepping frequency $f_{\mathrm{gait}}$, inter-leg timing offsets $(\theta_1,\theta_2,\theta_3)$, and stance duration $d$---primarily with velocity, while adjusting the realized body posture and stance geometry in response to terrain
changes.

\subsubsection{Cost of Transport}

Fig.~\ref{fig:cot_all}(a) presents the CoT of Unitree AlienGo as a function of commanded forward velocity on flat and rough terrains in simulation. The proposed HRL framework achieves consistent energy efficiency across the evaluated velocity range. CoT decreases from low speeds toward moderate velocities and remains stable at higher speeds, reflecting efficient dynamic locomotion. Despite terrain-induced disturbances, the proposed HRL framework maintains stable locomotion with limited CoT variation, demonstrating robustness of the hierarchical structure.

In addition, a direct comparison with established benchmark methods is performed. Fig.~\ref{fig:cot_all}(c) presents CoT results on flat terrain for the Unitree Go1 platform. The proposed HRL framework is compared against Walk These Ways (WTW)~\cite{margolis2023walk}, following the fixed trotting gait reported therein as the best gait in terms of energy consumption, the end-to-end approach Adaptive Energy Regularization (AER)~\cite{liang2025adaptive} with $\alpha=1$, and the hierarchical approach Fast \& Efficient (F\&E)~\cite{yang2022fast}, under identical evaluation conditions.
Across the mid-to-high velocity range, the proposed method achieves lower CoT compared to the three benchmarks. At lower speeds, performance remains better compared to both WTW and F\&E, while AER exhibits lower energy consumption in this regime. The interval-wise CoT results are summarized in Table~\ref{tab:cot_baselines}.




\subsubsection{Real-World Deployment}

\begin{figure*}[t]
\centering

\begin{minipage}[t]{0.32\textwidth}
    \centering

    \includegraphics[
        width=\linewidth,
        height=3.3cm,
        keepaspectratio
    ]{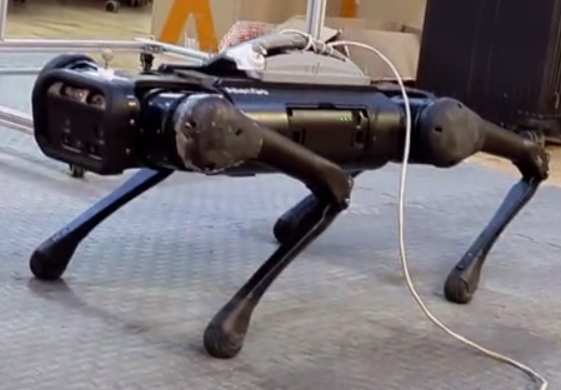}
\end{minipage}
\hfill
\begin{minipage}[t]{0.32\textwidth}
    \centering

    \includegraphics[
        width=\linewidth,
        height=3.3cm,
        keepaspectratio
    ]{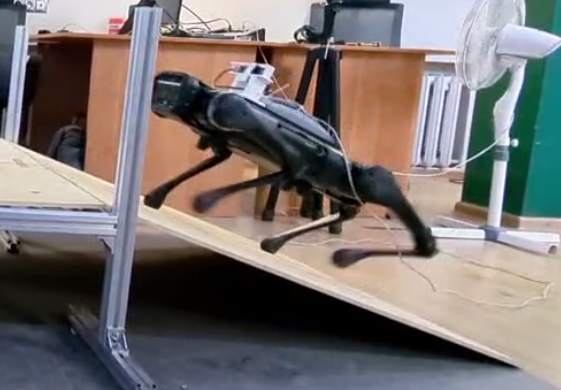}
\end{minipage}
\hfill
\begin{minipage}[t]{0.32\textwidth}
    \centering

    \includegraphics[
        width=\linewidth,
        height=3.3cm,
        keepaspectratio
    ]{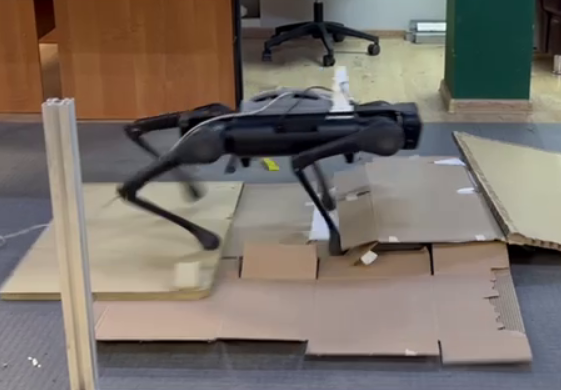}
\end{minipage}

\caption{Representative real-world locomotion trials on flat
terrain, inclined slope, and uneven rough terrain.}
\label{fig:real}
\end{figure*}

The trained HRL framework was deployed zero-shot on the physical Unitree AlienGo platform using the Unitree SDK and joint-space PD control with gains $K_p=50$ and $K_d=2$, without any additional finetuning on hardware. The gait transition--velocity relationship observed during deployment remained consistent with that in simulation, with smooth gait adaptation across commanded velocities. The robot also successfully traversed inclined terrain up to $20^\circ$, corresponding to the maximum training slope. In real-world robustness tests, the robot completed all 10 trials on flat, inclined, and uneven rough terrain without falls. Representative trials are shown in Fig.~\ref{fig:real}.


\begin{table*}[t]
    \centering
    \caption{AlienGo CoT ablation across commanded-velocity intervals.}
    \label{tab:aliengo_cot}

    \small
    \setlength{\tabcolsep}{5pt}
    \renewcommand{\arraystretch}{1.12}

    \resizebox{\textwidth}{!}{%
    \begin{tabular}{lcccc}
        \toprule
        \textbf{Method}
        & \textbf{(0,0.5]}
        & \textbf{(0.5,1.0]}
        & \textbf{(1.0,1.5]}
        & \textbf{(1.5,2.0)} \\
        \midrule

        HL w/ Energy, Train: Complex, Test: Flat
        & \meanstd{1.249}{0.486}
        & \meanstd{0.734}{0.069}
        & \meanstd{0.721}{0.037}
        & \meanstd{0.714}{0.024} \\

        HL w/ Energy, Train: Complex, Test: Rough
        & \meanstd{1.846}{1.333}
        & \meanstd{0.766}{0.074}
        & \meanstd{0.792}{0.054}
        & \meanstd{0.805}{0.034} \\

        HL w/ Energy, Train: Flat, Test: Rough
        & \meanstd{5.029}{4.149}
        & \meanstd{1.036}{0.178}
        & \meanstd{0.898}{0.052}
        & \meanstd{0.942}{0.182} \\

        HL w/o Energy, Train: Complex, Test: Flat
        & \meanstd{17.118}{8.179}
        & \meanstd{8.539}{0.317}
        & \meanstd{7.844}{0.277}
        & \meanstd{7.649}{0.405} \\

        HL w/o Energy, Train: Complex, Test: Rough
        & \meanstd{37.114}{37.478}
        & \meanstd{8.906}{0.348}
        & \meanstd{9.295}{1.615}
        & \meanstd{14.791}{5.193} \\

        \bottomrule
    \end{tabular}%
    }
\end{table*}

\begin{figure}[t]
    \centering
    \includegraphics[width=0.9\columnwidth]
    {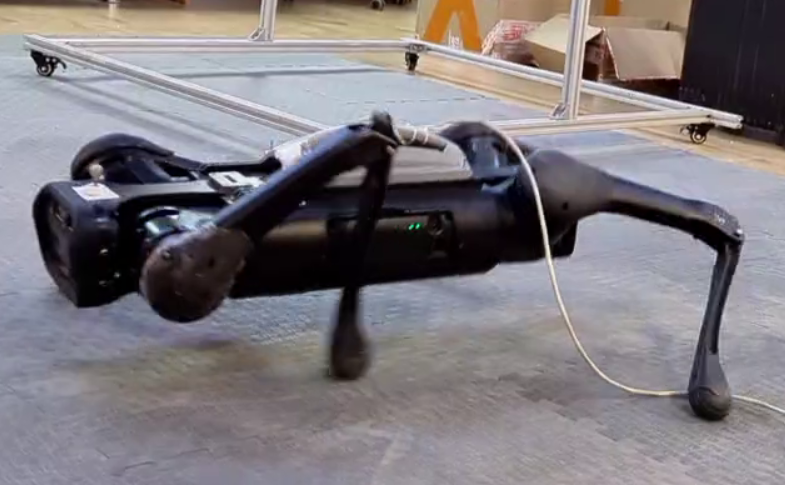}
    \caption{Qualitative reward-sensitivity case for an
    AER-style end-to-end policy on AlienGo with
    $\alpha=0.5$. Excessive reduction of joint effort resulted
    in a crouched and unstable posture.}
    \label{fig:aer_aliengo_failure}
\end{figure}

\begin{figure}[t]
    \centering
    \includegraphics[width=0.9\columnwidth]{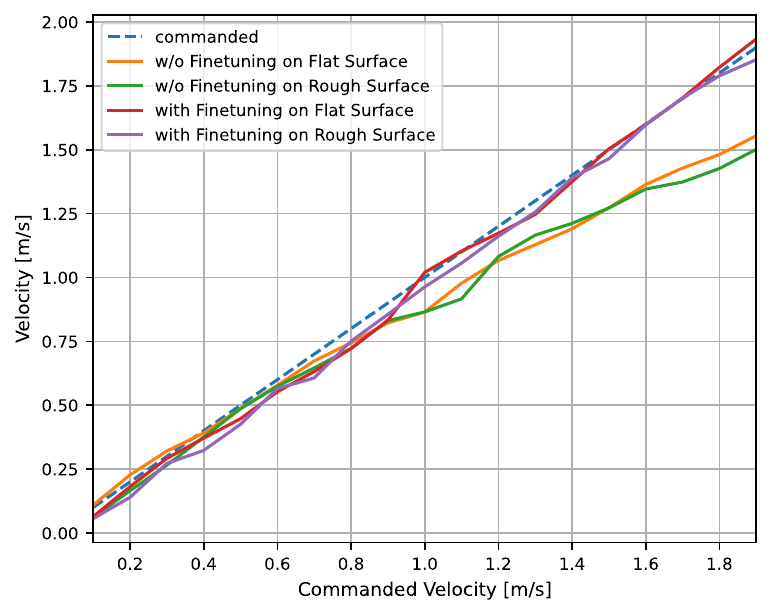}
    \caption{Velocity-tracking ablation for Unitree AlienGo. “w/o finetuning” denotes the LL policy obtained after Stage 2; “with finetuning” denotes the same policy after Stage 3. Both policies are evaluated on flat and rough terrain.}
    \label{fig:velocity_tracking}
\end{figure}

For qualitative comparison, we also deployed an AER-style
end-to-end policy on AlienGo, jointly optimizing the same
locomotion and energy objectives used in our framework~\cite{liang2025adaptive}. With
the energy reward weight set to $\alpha=0.5$, the policy reduced
joint effort excessively, resulting in the crouched posture shown
in Fig.~\ref{fig:aer_aliengo_failure}, where the generated torques
were insufficient to support the robot. While AER demonstrated
stable behavior on the smaller Go1 platform with $\alpha$ up to
$1.0$, this result illustrates the sensitivity of end-to-end energy
optimization to reward design and robot morphology. Our
HL formulation instead confines energy optimization to
interpretable gait parameters while preserving LL
stabilization.

\subsection{Ablation Study}

We evaluate the contributions of complex-terrain HL training,
the HL energy objective, and the third-stage LL finetuning.

\textit{High-Level Training Terrain:}
As shown in Fig.~\ref{fig:cot_all}(a) and
Table~\ref{tab:aliengo_cot}, the HL policy trained on complex
terrains maintains comparable CoT when deployed on flat and
rough surfaces. In contrast, training the HL policy only on flat
terrain increases CoT during zero-shot deployment on rough
terrain, particularly in the $(0,0.5]$\,m/s interval, where CoT
increases from $1.846$ to $5.029$. This demonstrates that
exposure to terrain variations during HL training improves the
generalization of HL gait adaptation.

\textit{High-Level Energy Reward:}
Removing the energy term substantially increases CoT across
all velocity intervals and both deployment terrains, as shown in
Fig.~\ref{fig:cot_all}(b) and
Table~\ref{tab:aliengo_cot}. This confirms that the accumulated
LL reward alone preserves locomotion but does not encourage
energy-efficient gait selection, whereas the explicit HL energy
objective is essential for reducing CoT.

\textit{Low-Level Finetuning:}
Fig.~\ref{fig:velocity_tracking} shows that the third training
stage improves velocity tracking on both flat and rough
terrains. The overall RMSE decreases from $0.165$ to $0.039$ m/s on
flat terrain and from $0.182$ to $0.047$ m/s on rough terrain. The improvement is most pronounced
above $1.0$ m/s. This improvement is not solely due to
adaptation to rough terrain. During the first stage, the LL
policy is trained using curriculum-sampled gait parameters,
whereas during deployment it receives the narrower and
correlated command distribution generated by the learned HL
policy. In the third stage, the HL policy is frozen and the LL
policy is optimized directly under these generated gait
commands.

\section{Discussion and Conclusion}

This work presented an HRL framework that separates energy-aware gait adaptation from high-frequency tracking and stabilization. It reduces average CoT by 27.7\%, 33.5\%, and 70.5\% compared with AER, WTW, and F\&E, respectively. The learned hierarchy exhibits automatic speed-dependent gait adaptation, transitioning from pacing at low speeds to trotting at higher speeds without predefined gait schedules or discrete switching, together with terrain-dependent adjustment of realized body posture and stance geometry. Simulation and real-world results demonstrate robust, energy-efficient locomotion across flat, rough, and inclined terrains. However, the framework remains limited by the low HL update frequency, which may reduce responsiveness to rapid terrain changes, and by the additional training cost and sensitivity to the decimation factor. Its performance also depends on the robustness of the LL controller, since execution and estimation errors may propagate through the hierarchy.

Future work will investigate adaptive HL update rates and decimation strategies, together with integrating vision-based perception into the proposed hierarchy. Exteroceptive terrain information may improve traversability in more challenging environments, including large irregular obstacles, stair-like structures, and steep slopes, by enabling the HL policy to exploit a broader portion of the gait-command space. However, expanding the scope of the HL policy may increase energy consumption. Therefore, an important objective is to determine whether visual terrain information can extend traversable environments while preserving the low CoT achieved by the current proprioceptive formulation.

\bibliographystyle{IEEEtran}
\bibliography{bibliography}

\end{document}